\documentclass{article}

\usepackage[preprint]{neurips_2021}

\usepackage[utf8]{inputenc} 
\usepackage[T1]{fontenc}    
\usepackage{url}            
\usepackage{booktabs}       
\usepackage{amsfonts}       
\usepackage{nicefrac}       
\usepackage{microtype}      
\usepackage{xcolor}         

\usepackage{amsmath}
\usepackage{amssymb}
\usepackage{bm}
\usepackage{graphicx}
\usepackage[pagebackref]{hyperref}
\usepackage{natbib}

\title{Look-ups are not (yet) all you need \\ for deep learning inference}

\author{%
  Calvin McCarter\thanks{Work done prior to employment at Amazon} \\
  \texttt{calmcc@amazon.com} \\
  \And
  Nicholas Dronen\textsuperscript{*}\\
  \texttt{ndronen@amazon.com} \\  
}

\begin{document}

\maketitle

\begin{abstract}
  Fast approximations to matrix multiplication have the potential to dramatically reduce the cost of neural network inference. Recent work on approximate matrix multiplication proposed to replace costly multiplications with table-lookups by fitting a fast hash function from training data. In this work, we propose improvements to this previous work, targeted to the deep learning inference setting, where one has access to both training data and fixed (already learned) model weight matrices.
  We further propose a fine-tuning procedure for accelerating entire neural networks while minimizing loss in accuracy.
  Finally, we analyze the proposed method on a simple image classification task.
  While we show improvements to prior work, overall classification accuracy remains substantially diminished compared to exact matrix multiplication.
  Our work, despite this negative result, points the way towards future efforts to accelerate inner products with fast nonlinear hashing methods. 
\end{abstract}

\section{Introduction}
To reduce the computational cost of dense matrix multiplications (or, more generally, inner products) in neural network (NN) inference, recent research efforts have sought to develop cheap approximations with minimal sacrifice of classification accuracy.
Methods based on distillation \citep{bucilua2006model,hinton2015distilling} and pruning \citep{han2015deep,janowsky1989pruning} have proposed to reduce the number of stored parameters and the number of multiply-accumulate (MAC) operations, while methods based on scalar quantization \citep{wu2016quantized} have proposed to reduce the size of stored parameters and the cost of each MAC.  
A recent method, \textsc{Maddness} \citep{blalock2021a}, instead proposed to approximate MACs by replacing them entirely with lookup-accumulate (LAC) operations. 
This reduced computational cost by involving fewer LACs (and less memory) than required by MACs, and by utilizing hardware multiplexers, which require fewer transistors than hardware multipliers.

\textsc{Maddness} breaks down a matrix multiply $\bm{A}\bm{B}$ (where $\bm{A}$ contains inputs / activations and $\bm{B}$ contains weights) into an assemblage of $\bm{a}^\top \bm{b}$ inner products, and approximates each of them by building on \citep{gray1984vector} and 
product quantization (PQ) \citep{gray1984vector,jegou2010product}. 
PQ enables fast inner products of the form $\bm{a}^\top \bm{b}$ when a training set of sampled $\bm{a}$s is given in advance. 
For paired vectors $\bm{a}$ and $\bm{b}$ of dimension $D$, the PQ algorithm partitions their dimensions to form $C \le D$ pairs of subvectors (and subspaces).
The inner product is then computed as the sum of approximate partial inner products over subspaces.
To do this, the $K$-Means algorithm is run over each subspace of the training set of $\bm{a}s$ to obtain $C$ subspace-specific sets of prototypes.
Since all $\bm{b}$s (corresponding to columns of NN weight matrices) are also known in advance, all possible inner products between subspace prototypes and $\bm{b}$ subvectors are precomputed offline and stored in $C$ lookup tables. 
(Each subspace-specific lookup table is referred to as a codebook, so $C$ is the number of codebooks.) 
To compute (approximate) inner products for a newly-seen $\bm{a}$ at inference time, PQ identifies, for each subspace, the nearest subspace prototype (represented by its index) for each subvector of $\bm{a}$.
This mapping from subvectors to categorical variables, represented as $\log_2(K)$-bit integer indices where $K$ is the number of prototypes per subspace, is called the \textit{encoding function}. 
Then, the computed inner product is simply the sum of partial inner products obtained from indexing into the appropriate rows in the precomputed lookup tables.

The primary contribution of \textsc{Maddness} addressed the computational bottleneck of PQ, with a fast new encoding function.
\textsc{Maddness} proposed a new hashing-based encoder, based on balanced binary regression trees, that runs efficiently on modern processors when $K=16$;
this encoder was learned from the training set of $\bm{a}s$ to minimize reconstruction error.
Also, rather than choosing prototypes via $K$-Means, \textsc{Maddness} optimized a least-squares objective, which also minimized reconstruction error; this objective conditioned on the training set of $\bm{a}$s, as well as the (already learned) encoder's resulting indices from the training samples. 
Finally, lookup tables were quantized to 8-bit block floating point format, with fast rounded 8-bit summation over subspaces.

While proposed to accelerate entire NNs, \textsc{Maddness} was validated for a large variety of models and settings but only on the final classification layers.
In each of these settings, the authors varied the number of codebooks, the key tuning parameter which trades off accuracy versus efficiency.
For a NN feedforward layer with $M$ output dimensions and \textsc{Float32} weights, the weight matrix requires $4DM$ bytes of storage.
With \textsc{Maddness}, the lookup table requires $KCM=16CM$ bytes of storage.

In this work, we propose a new method which builds on \textsc{Maddness}, and analyze its applicability to whole NN inference.
We dub our proposed method, Inference Targeted LookUp-based Matrix Multiplication (\textsc{ITLUMM}) \footnote{Spelled backward, this is MMULTI.}.
First, we intelligently partition the inner dimension rather than naively partitioning it into contiguous subspaces.
Second, we directly optimize the lookup table rather than prototypes, taking into account the subsequent nonlinear activation function (or classification loss function).
Furthermore, for accelerating entire neural networks, we propose incremental fine-tuning of linear layers.

We apply our approach to feedforward-only image classification on MNIST \citep{lecun2010mnist} and CIFAR-10 \citep{krizhevsky2009learning}.
We found that, while \textsc{Itlumm} improves upon \textsc{Maddness}, the accuracy-efficiency tradeoff is not yet good enough for practical use.
Our analysis of results suggests that an improved encoding function should be a key focus of future work.

\section{Method}

Here we describe \textsc{Itlumm}, which comprises intelligent subspace partitioning (Section \ref{sec:intelligent}), model-aware optimization of the lookup tables (Section \ref{sec:lut}), and fine-tuning of full networks (Section \ref{sec:finetuning}).

\subsection{Intelligent Subspace Partitioning}
\label{sec:intelligent}

For each subvector, \textsc{Maddness} performs exactly 4 comparisons in order to map the subvector into one of 16 lookup-table rows.
We would therefore prefer a subvector with maximal mutual information among its elements, such that this small number of comparisons provides maximal information about the entire subvector.
Thus, we should partition the vector $\bm{a}$ such that mutually-informative dimensions are placed together in the same subvector.
To do this, we find a permutation of the $D$ dimensions of $\bm{a}$, then slice the permuted dimensions into $C$ contiguous subspaces.

This problem is closely related to Optimized Product Quantization (OPQ) \citep{ge2013optimized} which rotates vectors before partitioning them, but in our case our rotation matrix must be a permutation matrix.
Directly optimizing for the optimal permutation is challenging, so we try two different approximate solutions for this problem. 

\subsubsection{OPQ-based Partitioning}

We first run OPQ to obtain a fully-dense rotation matrix $\bm{R}$. 
Then, we find a permutation matrix $\bm{Q}$ that approximates the effect of $\bm{R}$, by solving the following maximum weight matching problem:
\begin{gather}
    \max_{\bm{Q}} \sum_i \sum_j \bm{R}_{ij} \bm{Q}_{ij}.
\end{gather}
We solve this using the Hungarian algorithm \citep{kuhn1955hungarian}.
The resulting bipartite graph matches dimensions of the original vector to their ``closest matching'' dimensions in the OPQ-rotated vector.
Once we have a permuted ordering of the dimensions, we split the reordered dimensions into $C$ contiguous equally-sized chunks.

\subsubsection{Squared-correlation Hierarchical Clustering}

Here, we instead try to directly identify a permutation of the dimensions that places correlated dimensions closely together.
We first compute the squared correlation matrix $R^2$, then perform hierarchical/agglomerative clustering on it to produce a dendrogram of the dimensions \cite{mullner2013fastcluster}.
Next, we convert the dendrogram into a permutation by finding the ordering of the leaf nodes such that the distance between successive leaves is minimal.
Finally, we split the reordered dimensions into $C$ contiguous equally-sized chunks.

\subsection{Model-Aware Lookup-Table Optimization}
\label{sec:lut}

The prototype-optimization objective function of \textsc{Maddness} sought to maximize the ability to reconstruct $\bm{A}$, given the prototypes and the encoder's output (i.e. indices into the lookup-tables).
In other words, given a row vector $\bm{a}^\top$ and the corresponding encoder output $\bm{g}^\top$ (the concatenation of $C$ one-hot 16-dimensional vectors),
the prototype matrix $\bm{P}$ was optimized so that $\bm{a}^\top \approx \bm{g}^\top \bm{P}$.
To do this, given a training matrix $\bm{A}$, its corresponding encoder output matrix $\bm{G}$, and the original $K$-means prototype matrix $\bm{P}_0$, \textsc{Maddness} optimized the following:
\begin{gather}
    \arg\min_{\bm{P}} \|\bm{A} - \bm{GP}\|^2 + \lambda \|\bm{P}-\bm{P}_0\|^2,
\end{gather}
where $\lambda$ is a regularizer.
Finally, \textsc{Maddness} computed the lookup-table via $\bm{T} = \bm{P}\bm{B}$.

We note that this procedure ignores the fact that for deep learning inference, we also know the already-learned weight matrix $\bm{B}$.
We also observe that it aims at reconstructing the layer inputs $\bm{A}$, rather than the matrix product $\bm{AB}$.
In fact, in deep learning we actually care about not the matrix product $\bm{AB}$, but rather the layer output $\sigma(\bm{AB})$. The function $\sigma(\cdot)$ is the composition of the bias term with a nonlinearity (either an elementwise activation function such as ReLU or a row-wise nonlinearity for classification such as softmax).
Optimizing quantization for downstream performance has shown great success in previous works \cite{guo2020accelerating,may2019downstream}.

Therefore, in \textsc{Itlumm} we take advantage of our knowledge of both the model weights and the subsequent nonlinearity.
Furthermore, for computationally efficiency at training time, we directly optimize the lookup table, rather than the prototypes.
This leads to the following objective:
\begin{align}
    \arg\min_{\bm{T}} \|\sigma(\bm{AB}) - \sigma(\bm{GT})\|^2 + \lambda \|\bm{T}-\bm{P}_0\bm{B}\|^2.
\end{align}

\subsection{Fine-tuning for Acceleration of Full Neural Networks}
\label{sec:finetuning}

Replacing a neural network layer with approximate hashing-based lookups may degrade accuracy in two ways.
Layer replacement may alter the distribution of the layer's outputs without destroying information; inference accuracy would still be degraded since subsequent layers would undergo domain shift.
In this case, it would in principle be possible to retrain subsequent layers to adjust to this shift and regain lost accuracy, assuming subsequent layers are sufficiently high-capacity.
Alternatively, layer replacement may actually destroy information contained in the layer's inputs, and subsequent layers would be unable to recover original accuracy.

To accelerate full networks, while ameliorating the first cause of degraded accuracy, we propose incremental layer replacement.
For an $L$-layer NN with layers indexed as $l = 1,\dots,L$, we start with the input layer $1$ and proceed towards the classification layer $L$.
At each step $l$, we first replace the layer by collecting its inputs for the training data.
We then freeze layers $1,\dots,l$ (now using our non-differentiable lookups) and fine-tune the weights of layers $l+1,\dots,L$ (still using exact matrix multiplication) on the training data.

\section{Experiments}

\subsection{Approximating Softmax Classifiers}

We first repeat the softmax classifier experiments of \cite{blalock2021a} using the final dense classifier layers of VGG-like neural networks trained on CIFAR-100.
In both these networks, the final activations are 512-dimensional, so the $\bm{A}$ matrices are $10{,}000 \times 512$ in both cases, while the $\bm{B}$ matrices are $512 \times 10$ and $512 \times 100$, respectively.
Because MADDNESS was Pareto optimal compared to its prior works, we compare only it to our improved methods. 
Also, because the inference runtimes and memory usage are identical between \textsc{Maddness} and \textsc{Itlumm}, depending only on the number of codebooks, we show results in terms of number of codebooks rather than the runtime speedup (which relies on assumptions regarding hardware support and software implementation).
For each experimental setting, we plot the relative accuracy, the classification accuracy of a given setting divided by the accuracy obtained from using exact matrix multiplication.

\begin{figure}
    \centering
    \begin{tabular}{ccc}
    (A) & (B) & (C) \\
    \includegraphics[scale=0.47,clip=true,trim=4.2in 0 0in 0.3in]{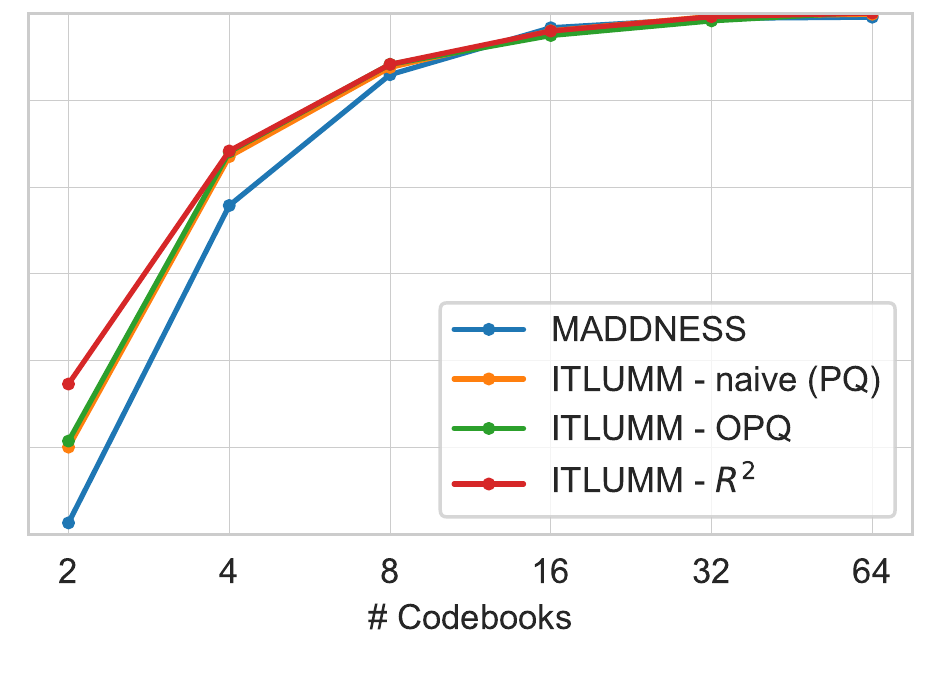} & \includegraphics[scale=0.47,clip=true,trim=4.2in 0 0in 0.3in]{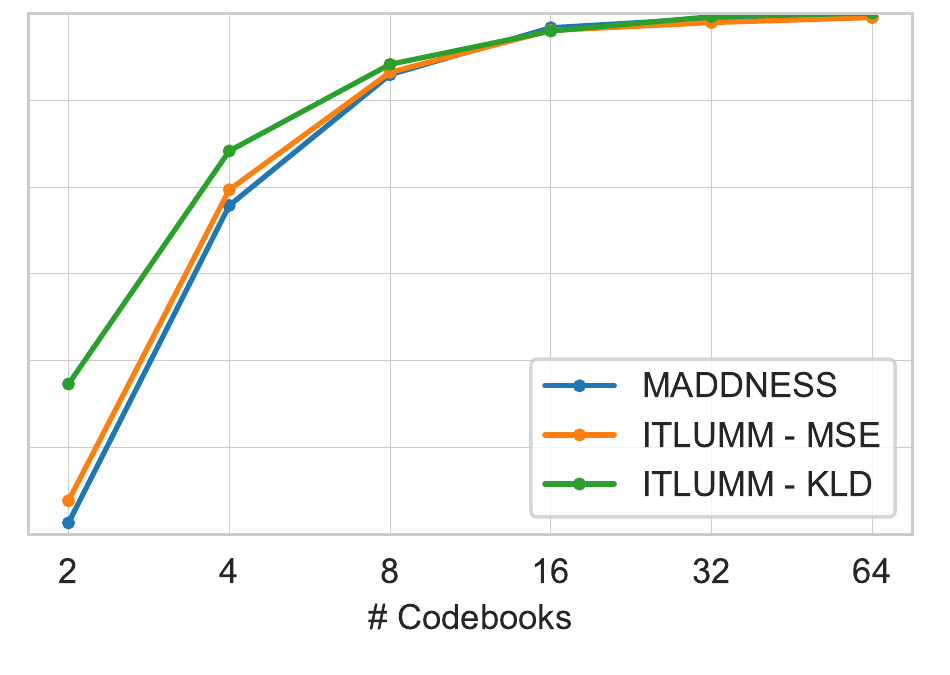} &
    \includegraphics[scale=0.47,clip=true,trim=4.2in 0 0in 0.3in]{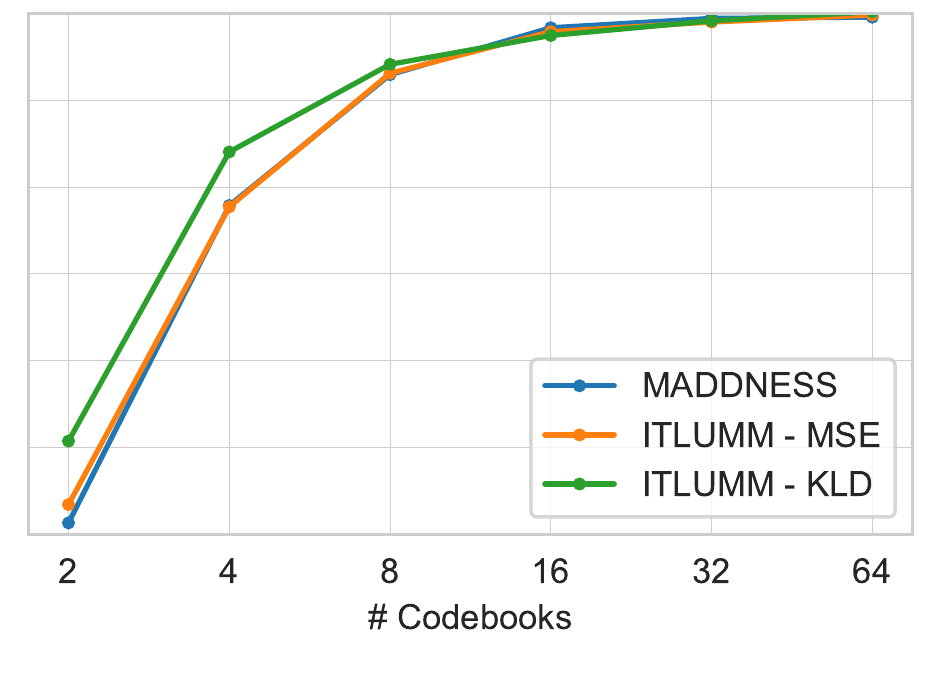}
    \end{tabular}
    \caption{Comparison of \textsc{Maddness} and \textsc{Itlumm} accuracy vs efficiency tradeoff for CIFAR-100 classifier layer. (A) Effect of \textsc{Itlumm} partitioning strategy.(B) Effect of nonlinearity function when \textsc{Itlumm} uses $R^2$-based partitioning.
    (C) Effect of nonlinearity function when \textsc{Itlumm} uses OPQ-based partitioning.}
    \label{fig:cifar100}
\end{figure}

Our results are shown in Figure \ref{fig:cifar100}.
Overall, we see that all settings of \textsc{Itlumm} (depicted in orange, green, and red) are superior to \textsc{Maddness} (depicted in blue).
We also see the effect of the partitioning heuristic in Figure \ref{fig:cifar100}(A).
For \textsc{Itlumm} we fix the prototype-optimization objective to be the KL-divergence; we also plot the accuracy of \textsc{Maddness} as a reference.
We see that the OPQ heuristic provides essentially no improvement over naive (PQ) partitioning. Meanwhile, $R^2$-clustering partitioning provides a modest improvement.
Meanwhile, we see the effect of the prototype-optimization objective in Figure \ref{fig:cifar100}(B \& C).
We see that, regardless of the choice of partitioning heuristic, minimizing the KL-divergence (KLD) is better than minimizing mean-squared error (MSE).

\subsection{Accelerating Full Networks}

Our goal is to replace all of a network's matrix multiplications with lookup tables. To that end, we first evaluated our proposed approach on a simple multi-layer perceptron (MLP) trained on the MNIST dataset. The network has 4 layers, with 30 neurons per layer. We began with an ablation study to measure the effect that substituting lookup for multiplication has on accuracy in each layer, then evaluated the effect of said substitution on every layer.

\begin{figure}[h!]
    \centering
    \includegraphics[width=4.0in]{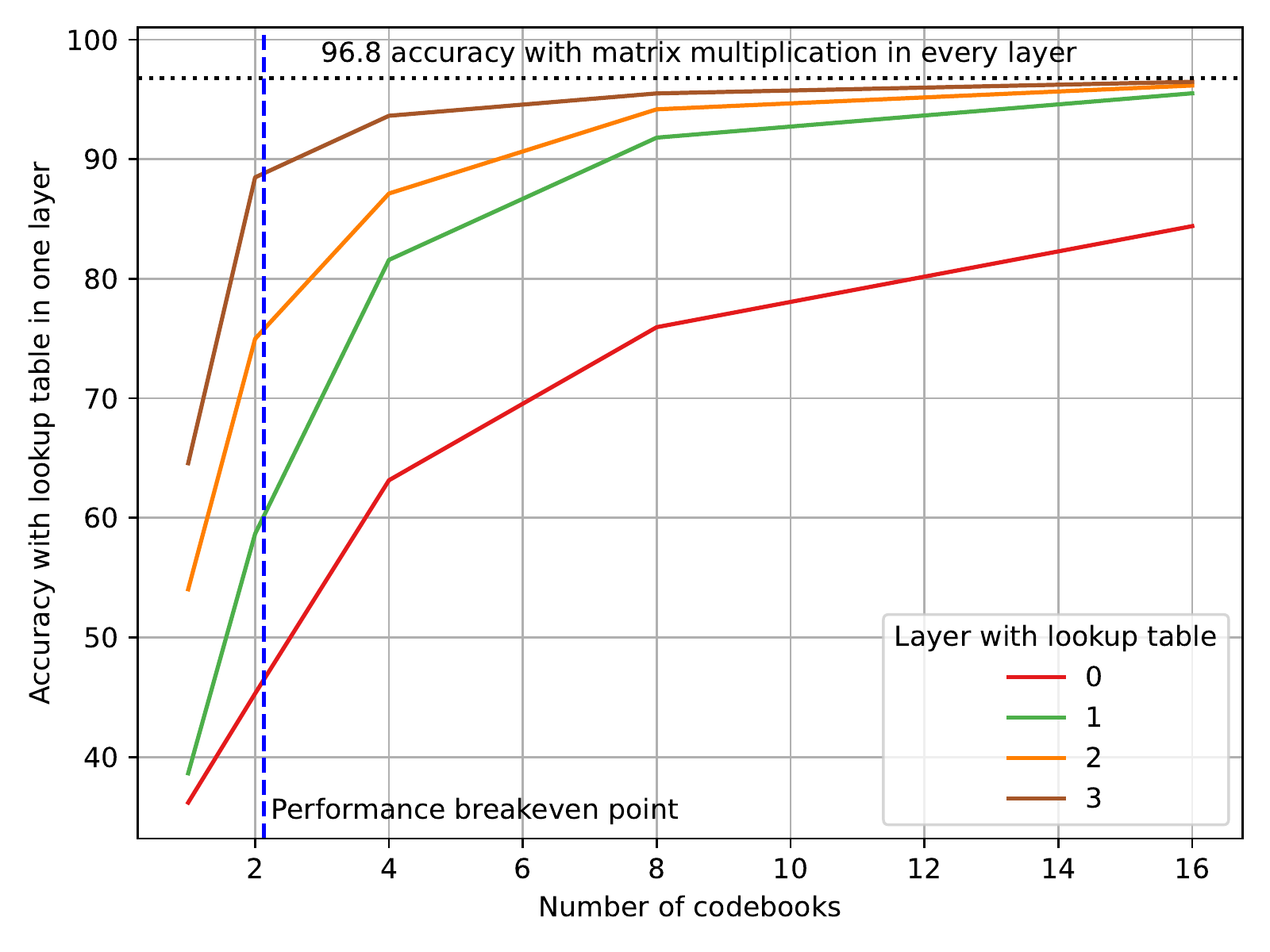}
    \caption{Effect of \textsc{Itlumm} on MNIST accuracy when converting a single layer to use \textsc{Itlumm} instead of matrix multiplication. The original network is a 4-layer MLP. The accuracy of the original network, with matrix multiplication in every layer, is the upper bound (black dotted line). Any configuration with fewer codebooks than the performance breakeven point is faster than matrix multiplication (blue dashed line).}
    \label{fig:mnist-ablation}
\end{figure}

Figure \ref{fig:mnist-ablation} shows the accuracy of a network as a function of the number of codebooks when a given layer's matrix multiplication has been replaced by a lookup table. Higher layers in the network, those closer to the output, are more robust to the error introduced by a lookup tables. However, the performance breakeven point is between 2 and 3 codebooks, and the degradation of accuracy is quite severe. Because of the compounding nature of error in neural networks, it's likely that replacing all of the layers in the network with \textsc{Itlumm} will result in even more severe degradation.

\begin{table}[]
    \centering
    \begin{tabular}{ccc}
    Number of codebooks & Accuracy & Faster \\
    \hline
    1 & 36.1 & \textsc{Yes} \\
    2 & 36.3 & \textsc{Yes} \\
    4 & 52.0 & \textsc{No} \\
    8 & 70.5 & \textsc{No} \\
    16 & 84.9 & \textsc{No} \\
    \end{tabular}
    \caption{Effect of \textsc{Itlumm} on MNIST accuracy when converting every layer to use \textsc{Itlumm} instead of matrix multiplication. A configuration in the table is faster than matrix multiplication when its number of codebooks is less than the breakeven point in Figure \ref{fig:mnist-ablation}.}
    \label{tab:mnist-full-network}
\end{table}

Indeed, Table \ref{tab:mnist-full-network} shows that replacing all of the matrix multiplications in this network with \textsc{Itlumm} results in accuracy worse than that due to replacing any single layer. Since the performance breakeven point is 2 codebooks, accelerating the entire network \textsc{Itlumm} requires a loss of more than 60\% accuracy.

\section{Discussion}

As expected, \textsc{ITLUMM}'s model-aware lookup-table optimization improves upon \textsc{Maddness}. 
Meanwhile, optimized partitioning provides little improvement, surprisingly so given the substantial benefits of OPQ in maximum inner-product search (MIPS) \citep{norouzi2013cartesian,ge2013optimized}.
However, we are not able to obtain the full benefits of OPQ because the applied rotation must be a permutation.
Furthermore, in the MIPS setting, input data are expected to have redundancy/correlation and a wide range of variances,
while NN activations are regularized (e.g. with dropout) \cite{srivastava2014dropout} to avoid this.

We suggest that future research focus on improving the \textsc{Maddness} hash function, which appears to be an accuracy bottleneck.
One could exploit knowledge of weight matrix $\bm{B}$ while learning hash function parameters, as done for NN scalar quantization \cite{stock2020and}.
More substantially, a differentiable hash function could improve inference accuracy and speed up conversion of full networks.

\section{Conclusion}

We proposed \textsc{Itlumm}, an approach for accelerating matrix multiplication by replacing multiplications with look-ups, which improves upon previous work.
In the context of deep learning inference, where model weights are known, our approach advanced classification accuracy beyond prior work in \textsc{Maddness} \citep{blalock2021a}.
Our approach and analysis on full neural networks informs the community on the current state of hashing-based acceleration, and points the way to potential future advances.
\clearpage
\bibliographystyle{authordate1}
\bibliography{main}

\begin{thebibliography}{}

\bibitem[\protect\citename{Blalock \& Guttag, }2021]{blalock2021a}
Blalock, Davis, \& Guttag, John. 2021.
\newblock Multiplying Matrices Without Multiplying.
\newblock {\em Pages  992--1004 of:} Meila, Marina, \& Zhang, Tong (eds), {\em
  Proceedings of the 38th International Conference on Machine Learning}.
\newblock Proceedings of Machine Learning Research, vol. 139.
\newblock PMLR.

\bibitem[\protect\citename{Bucilua {\em et~al.}, }2006]{bucilua2006model}
Bucilua, Cristian, Caruana, Rich, \& Niculescu-Mizil, Alexandru. 2006.
\newblock Model compression.
\newblock {\em Pages  535--541 of:} {\em Proceedings of the 12th ACM SIGKDD
  international conference on Knowledge discovery and data mining}.

\bibitem[\protect\citename{Ge {\em et~al.}, }2013]{ge2013optimized}
Ge, Tiezheng, He, Kaiming, Ke, Qifa, \& Sun, Jian. 2013.
\newblock Optimized product quantization.
\newblock {\em IEEE transactions on pattern analysis and machine intelligence},
  {\bf 36}(4), 744--755.

\bibitem[\protect\citename{Gray, }1984]{gray1984vector}
Gray, Robert. 1984.
\newblock Vector quantization.
\newblock {\em IEEE Assp Magazine}, {\bf 1}(2), 4--29.

\bibitem[\protect\citename{Guo {\em et~al.}, }2020]{guo2020accelerating}
Guo, Ruiqi, Sun, Philip, Lindgren, Erik, Geng, Quan, Simcha, David, Chern,
  Felix, \& Kumar, Sanjiv. 2020.
\newblock Accelerating large-scale inference with anisotropic vector
  quantization.
\newblock {\em Pages  3887--3896 of:} {\em International Conference on Machine
  Learning}.
\newblock PMLR.

\bibitem[\protect\citename{Han {\em et~al.}, }2015]{han2015deep}
Han, Song, Mao, Huizi, \& Dally, William~J. 2015.
\newblock Deep compression: Compressing deep neural networks with pruning,
  trained quantization and huffman coding.
\newblock {\em arXiv preprint arXiv:1510.00149}.

\bibitem[\protect\citename{Hinton {\em et~al.}, }2015]{hinton2015distilling}
Hinton, Geoffrey, Vinyals, Oriol, Dean, Jeff, {\em et~al.} 2015.
\newblock Distilling the knowledge in a neural network.
\newblock {\em arXiv preprint arXiv:1503.02531}, {\bf 2}(7).

\bibitem[\protect\citename{Janowsky, }1989]{janowsky1989pruning}
Janowsky, Steven~A. 1989.
\newblock Pruning versus clipping in neural networks.
\newblock {\em Physical Review A}, {\bf 39}(12), 6600.

\bibitem[\protect\citename{Jegou {\em et~al.}, }2010]{jegou2010product}
Jegou, Herve, Douze, Matthijs, \& Schmid, Cordelia. 2010.
\newblock Product quantization for nearest neighbor search.
\newblock {\em IEEE transactions on pattern analysis and machine intelligence},
  {\bf 33}(1), 117--128.

\bibitem[\protect\citename{Krizhevsky {\em et~al.},
  }2009]{krizhevsky2009learning}
Krizhevsky, Alex, Hinton, Geoffrey, {\em et~al.} 2009.
\newblock Learning multiple layers of features from tiny images.

\bibitem[\protect\citename{Kuhn, }1955]{kuhn1955hungarian}
Kuhn, Harold~W. 1955.
\newblock The Hungarian method for the assignment problem.
\newblock {\em Naval research logistics quarterly}, {\bf 2}(1-2), 83--97.

\bibitem[\protect\citename{LeCun {\em et~al.}, }2010]{lecun2010mnist}
LeCun, Yann, Cortes, Corinna, \& Burges, Christopher~J. 2010.
\newblock MNIST handwritten digit database. 2010.
\newblock {\em URL http://yann. lecun. com/exdb/mnist}, {\bf 7}(23), 6.

\bibitem[\protect\citename{May {\em et~al.}, }2019]{may2019downstream}
May, Avner, Zhang, Jian, Dao, Tri, \& R{\'e}, Christopher. 2019.
\newblock On the downstream performance of compressed word embeddings.
\newblock {\em Advances in neural information processing systems}, {\bf 32}.

\bibitem[\protect\citename{M{\"u}llner, }2013]{mullner2013fastcluster}
M{\"u}llner, Daniel. 2013.
\newblock fastcluster: Fast hierarchical, agglomerative clustering routines for
  R and Python.
\newblock {\em Journal of Statistical Software}, {\bf 53}, 1--18.

\bibitem[\protect\citename{Norouzi \& Fleet, }2013]{norouzi2013cartesian}
Norouzi, Mohammad, \& Fleet, David~J. 2013.
\newblock Cartesian k-means.
\newblock {\em Pages  3017--3024 of:} {\em Proceedings of the IEEE Conference
  on computer Vision and Pattern Recognition}.

\bibitem[\protect\citename{Srivastava {\em et~al.},
  }2014]{srivastava2014dropout}
Srivastava, Nitish, Hinton, Geoffrey, Krizhevsky, Alex, Sutskever, Ilya, \&
  Salakhutdinov, Ruslan. 2014.
\newblock Dropout: a simple way to prevent neural networks from overfitting.
\newblock {\em The journal of machine learning research}, {\bf 15}(1),
  1929--1958.

\bibitem[\protect\citename{Stock {\em et~al.}, }2020]{stock2020and}
Stock, Pierre, Joulin, Armand, Gribonval, R{\'e}mi, Graham, Benjamin, \&
  J{\'e}gou, Herv{\'e}. 2020.
\newblock And the Bit Goes Down: Revisiting the Quantization of Neural
  Networks.
\newblock {\em Pages  1--11 of:} {\em ICLR 2020-Eighth International Conference
  on Learning Representations}.

\bibitem[\protect\citename{Wu {\em et~al.}, }2016]{wu2016quantized}
Wu, Jiaxiang, Leng, Cong, Wang, Yuhang, Hu, Qinghao, \& Cheng, Jian. 2016.
\newblock Quantized convolutional neural networks for mobile devices.
\newblock {\em Pages  4820--4828 of:} {\em Proceedings of the IEEE conference
  on computer vision and pattern recognition}.

\end{thebibliography}

\end{document}